\pdfoutput=1

\documentclass[a4paper,fleqn=false]{cas-sc}

\usepackage[numbers]{natbib}
\usepackage{tabularx}
\usepackage{array}
\usepackage{multirow}
\usepackage{booktabs}
\usepackage{enumitem}

\def\tsc#1{\csdef{#1}{\textsc{\lowercase{#1}}\xspace}}
\tsc{WGM}
\tsc{QE}
\tsc{EP}
\tsc{PMS}
\tsc{BEC}
\tsc{DE}

\begin{document}
\let\WriteBookmarks\relax
\def\floatpagepagefraction{1}
\def\textpagefraction{.001}
\shorttitle{Generalized Synthetic Image Detection with Enhanced RGB-Noise Representation Learning}
\shortauthors{Zhen Li et~al.}

\title [mode = title]{Generalized Synthetic Image Detection with Enhanced RGB-Noise Representation Learning}                      

\author[1]{Zhen Li}[
	style=chinese,
    orcid=0009-0009-9888-4513,
]
\ead{lizhen@mails.cuc.edu.cn}

\author[1,2]{Gang Cao}[
	style=chinese,
    orcid=0000-0002-4549-0125,
]
\ead{gangcao@cuc.edu.cn}
\cormark[1]

\author[1]{Tian Zhang}[
	style=chinese,
    orcid=0009-0007-0351-3206
]
\ead{zt0226@mails.cuc.edu.cn}

\author[3]{Lifang Yu}[
	style=chinese,
    orcid=0000-0002-0508-7526,
]
\ead{yulifang@bigc.edu.cn}

\author[4]{Shaowei Weng}[
	style=chinese,
    orcid=0000-0003-1037-7699,
]
\ead{wswweiwei@126.com}

\affiliation[1]{organization={School of Computer and Cyber Sciences, Communication University of China},
	city={Beijing},
	postcode={100024},
	country={China}}    

\affiliation[2]{organization={School of Information Engineering, Changsha Medical University},
	city={Changsha},
	postcode={410219},
	country={China}}

\affiliation[3]{organization={Department of Information Engineering, Beijing
Institute of Graphic Communication},
	city={Beijing},
	postcode={100026},
	country={China}}

\affiliation[4]{organization={Fujian Provincial Key Laboratory of Big Data
Mining and Applications, Fujian University of Technology},
	city={Fuzhou},
	postcode={350118},
	country={China}}

\cortext[cor1]{Corresponding author}

\begin{abstract}
The rapid advancement of large-scale generative models has accelerated the spread of highly deceptive AI-generated images, making generalized synthetic image detection a critical imperative. Existing forensic networks often struggle with cross-model generalization and real-world degradations due to their reliance on single-domain representations and conventional binary classification optimization. To overcome these limitations, we propose RNSIDNet, a novel forensic framework that achieves robust detection through enhanced RGB-Noise representation learning. Specifically, our method employs a dual-branch architecture where global RGB semantics, extracted by an attention-refined CLIP backbone, dynamically modulate high-frequency noise artifacts captured by Bayar convolutions via a Feature-wise Linear Modulation (FiLM) module. To further enhance the learned representations, we design a Hard Sample-aware Contrastive Learning (HSCL) strategy. By explicitly penalizing challenging training samples, HSCL reshapes the latent feature space to maximize the discriminative margin between pristine and synthetic domains. Extensive experiments across eight public benchmark datasets verify that our model achieves state-of-the-art performance, delivering superior generalization ability, robustness, and computational efficiency. Code and dataset will be publicly available on \url{https://github.com/multimediaFor/RNSIDNet}.
\end{abstract}




\maketitle
\section{Introduction}

The rapid evolution of deep learning techniques, especially large-scale generative models~\citep{goodfellow2014generative, ho2020denoising, rombach2022high, podell2024sdxl}, has significantly advanced image synthesis while simultaneously facilitating the spread of deceptive AIGC content and deepfakes. In response, researchers have developed various forensic methods to identify these forgeries within a learning-based framework~\citep{wang2020cnn,ojha2023towards,bai2024ai,lou2025exploring}. 

While numerous detection methods have been proposed, they frequently encounter a dual bottleneck in feature representation and model optimization when facing unseen generative models or complex real-world degradations, limiting their real-world applicability. A primary limitation of existing detectors is their reliance on single-domain representation. Spatial models are highly prone to overfitting dataset-specific semantics, whereas frequency-based approaches degrade sharply under common image corruptions like JPEG compression or Gaussian blur~\citep{wang2020cnn, frank2020leveraging}. To capture comprehensive forensic traces, recent studies~\citep{yan2025a,tan2024rethinking} have attempted to fuse diverse feature representations. Unfortunately, most existing methods rely on naive fusion strategies, such as direct concatenation or element-wise addition. These rigid operations fail to capture the complex contextual dependencies between different modalities. Consequently, such ineffective integration ultimately forces heterogeneous features to interfere with each other rather than achieve synergy, where dominant semantic information may obscure subtle frequency-domain artifacts.

Beyond representation, traditional optimization objectives further constrain detection capabilities. Most detectors~\citep{wang2020cnn,ojha2023towards,wang2023dire} frame synthetic image detection as a standard binary classification task supervised by a Binary Cross-Entropy (BCE) loss. However, it creates a vulnerable decision boundary that is easily bypassed by the high-quality outputs of modern AIGC models. Previous works~\citep{chen2024drct,koutlis2024leveraging} adopt contrastive learning to improve the feature representation, but they treat all sample pairs uniformly. For deepfake detection, most mismatched pairs which are visually distinct from the anchor sample can be easily pushed away, while the most discriminative information comes from a small set of hard samples that lie extremely close to the anchor in feature space. In this way, the contribution of these critical samples is diluted by the overwhelming number of easy ones.

To tackle these challenges, we propose an RGB-Noise dual-branch Synthetic Image Detection Network (RNSIDNet). It aims to construct a robust framework for generalized synthetic image detection, utilizing enhanced RGB-Noise representation learning to deeply synergize heterogeneous features and reshape the representation space. At the architectural level, we design a gating-based dynamic feature fusion mechanism. Instead of conventional direct concatenation, it innovatively utilizes the RGB features as conditioning variables to dynamically generate scale and bias parameters via Feature-wise Linear Modulation (FiLM)~\citep{perez2018film}, which subsequently modulate the noise branch features. Furthermore, we introduce a Hard Sample-aware Contrastive Learning (HSCL) strategy. It achieves dynamic re-weighting within a compact low-dimensional manifold, actively tracking and assigning larger gradient penalties to highly deceptive images. This spatial optimization effectively strips away gradient interference from simple samples, significantly enhancing intra-class clustering and inter-class repulsion, thereby successfully widening and reshaping the decision boundary against extremely photorealistic forgeries. Finally, building upon existing research, we construct and release a large-scale Aligned Multi-source Synthetic Image Dataset (AMSID). By enforcing strict pixel alignment between pristine and synthetic sample pairs across diverse generators, AMSID enables the network to suppress content-driven distribution biases and focus on intrinsic generative artifacts.

Through such specialized designs, RNSIDNet achieves high generalization capabilities across various generative models and diverse content types. Notably, even when trained on limited data and with a mere fraction of the trainable parameters, RNSIDNet yields performance comparable to massive models trained on substantially larger datasets, demonstrating exceptional efficiency and robustness.

In summary, the main contributions of this paper are as follows:

\begin{itemize}
    \item We propose RNSIDNet, an RGB-noise synthetic image detection network, by dynamically modulating noise features with RGB modal, achieving an effective fusion of heterogeneous representations.
    \item We propose a representation enhancement strategy guided by a Hard Sample-aware Contrastive Loss (HSCL). It forces the feature space to achieve higher inter-class separability and intra-class compactness by amplifying the repulsion against hard cross-class samples.
    \item Extensive experiments on eight public datasets demonstrate that the proposed method achieves the state-of-the-art detection performance, exhibiting outstanding generalization capabilities in cross-model forgery detection.
\end{itemize} 

The rest of this paper is organized as follows. Section II reviews previous related works. The proposed RNSIDNet network and its training are elaborated in Section III. The experimental results and discussion are depicted in Section IV, followed by the conclusion drawn in Section V.

\section{Related Works}
\subsection{Single-Domain Feature}
Most end-to-end synthetic image detection methods focus on capturing visual artifacts or statistical anomalies within a single modality of the feature domain. In the \textbf{spatial domain}, researchers typically utilize CNNs to automatically learn generalizable pixel-level discriminative features. For instance, \citet{wang2020cnn} systematically verifies the existence of cross-GAN fingerprints. With the rise of large vision-language models, the powerful representation capability of pre-trained CLIP~\citep{radford2021learning} models is used to enhance cross-model detection generalization~\citep{ojha2023towards,cozzolino2024raising}. The subsequent RINE method~\citep{koutlis2024leveraging} leverages the intermediate features of CLIP encoder blocks. In the \textbf{frequency domain}, spectral analysis provides another perspective for image forensics. Early works~\citep{zhang2019making, frank2020leveraging} rely on basic transforms (DFT, DCT) to expose spectrum replication. Recent methods capture subtle generative artifacts more efficiently through diverse schemes, like frequency-aware networks~\citep{tan2024rethinking}, and hybrid approaches that extract noise residuals via a pre-trained denoising model or specialized filter to amplify high-frequency anomalies~\citep{liu2022detecting,lou2025trusted,zhao2022hybrid,guo2024effective}.

\subsection{Multi-Dimensional Feature Fusion}
To mitigate the inherent vulnerabilities and domain-specific overfitting of single-modality representations, multi-dimensional fusion frameworks have emerged as a promising paradigm. By concurrently capturing high-level semantics and low-level statistical artifacts, these frameworks significantly bolster detection resilience and generalization. In practice, some recent methodologies~\citep{ju2022fusing} leverage attention mechanisms to dynamically integrate global semantics with fine-grained local patches, while others exploit cross-modal alignments between textual prompts and visual content~\citep{sha2023fake,cozzolino2024raising}. Furthermore, advanced approaches synergize multi-domain cues—such as frequency spectra and local textures—with the robust representational priors of vision-language models~\citep{yan2025a,liu2024forgeryaware}.

\subsection{Training Paradigm and Optimization Strategy}
Recently, image reconstruction has emerged as a powerful proxy for exposing generative artifacts by measuring pixel-level reconstruction discrepancies. To mitigate the "generalization illusion" caused by overfitting to dataset-specific biases~\citep{guillaro2025bias}, recent studies align real and synthetic data distributions via autoencoders or self-conditioned reconstruction~\citep{rajan2025aligned, wang2023dire, chen2025dual}. Moreover, to overcome the rigid decision boundaries of traditional BCE loss, contrastive learning is increasingly integrated to directly enforce intra-class compactness and inter-class separability. Recent methods tailor this contrastive objective to specific feature levels, for instance, \citet{chen2024drct} construct contrastive pairs by widening low-level reconstruction trajectories, whereas \citet{tan2025c2p} apply contrastive optimization within the high-level semantic space using prompt-guided CLIP.

\section{Proposed scheme}
In this section, the architecture design, training dataset generation and training strategy of our proposed synthetic image detection network, i.e., RNSIDNet, are presented in detail.

\begin{figure}
    \centering
    \includegraphics[width=1\linewidth]{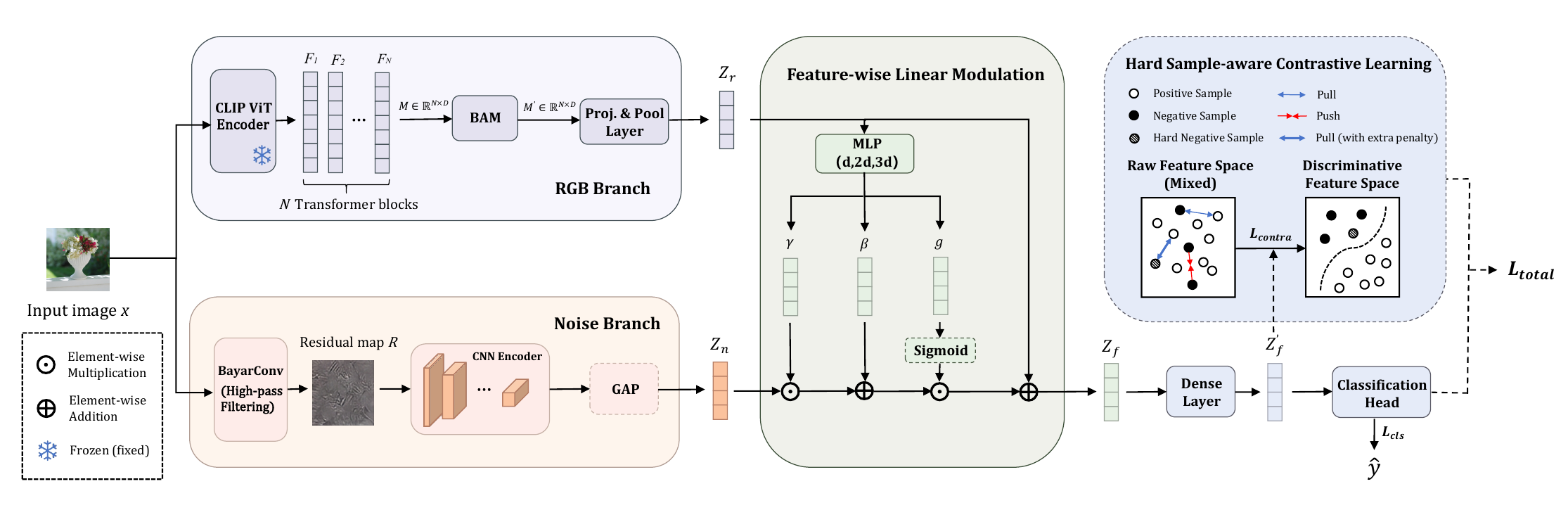}
    \caption{Overview of the proposed RNSIDNet framework. The input image is processed through two parallel branches: an RGB branch that extracts multi-scale representations using a frozen CLIP-ViT encoder and a Balanced Attention Module (BAM), and a noise branch that captures high-frequency residuals via Bayar convolution followed by Global Average Pooling (GAP). The heterogeneous features are subsequently integrated using a dynamic Feature-wise Linear Modulation (FiLM) module guided by the RGB features. Finally, the network is jointly optimized by a classification loss ($L_{cls}$) and a contrastive loss ($L_{contra}$) driven by the Hard Sample-aware Contrastive Learning (HSCL) strategy, explicitly enforcing intra-class compactness and inter-class separation.}
    \label{fig:architecture}
\end{figure}

\subsection{Network architecture}
\subsubsection{Overview}
The overall architecture of the proposed RNSIDNet is illustrated in Figure \ref{fig:architecture}. It employs a collaborative dual-branch structure introducing an RGB branch and a noise branch. The blind detection of AI-generated images is achieved by fusing high-level spatial representations with low-level noise residual signals. 

Given an input image $X \in \mathbb{R}^{H \times W \times 3}$, the RGB branch extracts the multi-scale spatial feature $M$ from a frozen CLIP-ViT~\citep{radford2021learning} encoder, which are further refined by a Balanced Attention Module (BAM)~\citep{park2018bam} to yield the RGB representation $Z_r$. Concurrently, the Noise Branch extracts high-frequency residual signals via a Bayar-constrained filter, followed by a lightweight CNN to produce the noise representation $Z_n$. During the feature fusion stage, we design a dual-branch fusion mechanism based on Feature-wise Linear Modulation (FiLM)~\citep{perez2018film}. This module leverages the extracted RGB image feature $Z_r$ as conditioning guidance to dynamically weight and filter the noise feature $Z_n$. Finally, $Z_f$ is passed through a classification head to output the forgery probability of the image, thereby completing the end-to-end detection task.

\subsubsection{Attention-Refined Spatial Feature Extraction}
In the RGB branch, the input image $X$ is fed into the frozen CLIP-ViT encoder. The layer-normalized output feature $F_i \in \mathbb{R}^D$ is extracted from the $i$-th Transformer~\citep{vaswani2017attention} block. Thereafter, these intermediate features are stacked along the layer dimension to construct a cross-layer feature representation $M \in \mathbb{R}^{N \times D}$, where $N$ denotes the number of extracted blocks. Such a representation successfully captures hierarchical semantic information across various network depths. Given that different layers contribute unequally to forgery detection, an adaptive re-weighting mechanism is designed to highlight salient cues.

To this end, a BAM module is employed to enhance the feature representation $M$. Specifically, the channel attention branch compresses the layer dimension to aggregate global context. The intermediate representation is processed by a shared Multi-Layer Perceptron (MLP) to derive the channel-wise weight vector $w_c \in \mathbb{R}^{1 \times D}$:
\begin{equation}
w_c = \sigma\left(\text{MLP}(\text{AvgPool}(M)) + \text{MLP}(\text{MaxPool}(M))\right)
\end{equation}
where $\sigma(\cdot)$ denotes the sigmoid activation function, and pooling operations are performed along the layer dimension $N$. Concurrently, the layer-wise attention branch captures the scaling factors across different network depths. It aggregates the feature dimension via mean and max pooling, and applies a 1D convolution to generate the layer-wise weight vector $w_l \in \mathbb{R}^{N \times 1}$:
\begin{equation}
w_l = \sigma\left(\text{Conv1d}([\text{MeanPool}(M) ; \text{MaxPool}(M)])\right)
\end{equation}
where $[\cdot ; \cdot]$ represents the concatenation along the feature dimension. Subsequently, $w_c$ and $w_l$ are broadcasted to $\mathbb{R}^{N \times D}$ by expanding along the layer and feature dimensions, respectively. They are then summed and activated by another sigmoid function to construct the joint attention map. Through a residual connection, the intermediate refined feature representation $\tilde{M}$ is obtained:
\begin{equation}
\tilde{M} = M + M \odot \sigma(w_c + w_l)
\end{equation}
where $\odot$ represents element-wise multiplication. Ultimately, to preserve the stability of the original semantic structure during training, a learnable gating parameter $\gamma$ is introduced to dynamically blend the original and intermediate features, yielding the final representation $M'$:
\begin{equation}
M' = (1 - \sigma(\gamma)) M + \sigma(\gamma) \tilde{M}.
\end{equation}
Following this gating mechanism, $M'$ is transformed into a $d$-dimensional vector by a projection layer and processed by a hybrid pooling layer. The final RGB feature vector $Z_r \in \mathbb{R}^d$ is computed as the mean of max and average-pooled representations.

\subsubsection{Constrained High-Frequency Noise Residual Learning}
In the noise branch, a constrained convolutional layer is utilized to extract high-frequency residual from the input image $X$. Existing methods typically rely on standard CNNs or Spatial Rich Model (SRM)~\citep{fridrich2012rich} filters, both of which have notable limits. Driven by the inherent optimization bias to minimize loss effortlessly, standard CNNs tend to greedily learn dominant low-frequency semantics, often degenerating into mere "content extractors" and causing severe feature homogenization with the RGB branch. Conversely, the hand-crafted SRM filter employs fixed kernels, lacking the dynamic adaptability required to capture the novel and evolving microscopic generative artifacts incurred by modern AIGC models. 

To overcome these bottlenecks, we introduce Bayar convolution~\citep{bayar2016deep}, which restricts the parameter space through structural constraints, guaranteeing that the filter maintains stable high-pass characteristics. Let the convolutional kernel size be $K \times K$, with its center coordinate denoted as $c$. This filter updates the gradients only for the $K^2 - 1$ non-center parameters, defined as $\theta = \{w_i \mid i\neq c\}$. During the forward propagation phase, the network first normalizes the non-center parameters, forcing them to satisfy $\sum_{i \neq c} w_i = 1$. Subsequently, a fixed weight $w_c = -1$ is inserted at the center of the kernel, constructing the complete convolutional kernel $W = \{w_1, \dots, w_{c-1}, -1, w_{c+1}, \dots, w_{K^2}\}$. Since the sum of the non-center weights is 1 and the center weight is fixed at -1, the convolutional kernel strictly satisfies a zero-sum constraint, i.e., $\sum_{i=1}^{K^2} w_i = 0$. This zero-sum architecture renders the convolutional kernel as a learnable high-pass filter. It captures the difference between the current pixel and the weighted average of its local neighbors. Consequently, the low-frequency semantic structural information of the image is effectively suppressed, while the local statistical anomaly and artifact residual are amplified. When dealing with multi-channel inputs, this constrained convolution operates independently on each input channel and is subsequently recombined along the output channel dimension, realizing cross-channel residual signal modeling. Let $\mathcal{B}(\cdot)$ be the constrained filtering operation defined by $W$, then the extracted residual response map is yielded as
\begin{equation}
 R = \mathcal{B}(X). 
\end{equation}
The resulting residual map $R$ contains high-frequency information, such as interpolation artifacts, compression traces, and the local textural anomaly unique to generative models. Subsequently, $R$ is fed into a lightweight convolutional encoding network $f_n(\cdot)$, which consists of five stacked convolutional blocks. Each block sequentially applies a $3 \times 3$ convolution with stride $2$, batch normalization, and a ReLU activation function, while doubling the channel dimension at each stage. Finally, a flatten operation compresses the spatial dimensions to yield the fixed-length noise feature representation $Z_n \in \mathbb{R}^d$ as
\begin{equation}
 Z_n = f_n(R)
\end{equation}
where $d$ is the final dimension of the flattened feature. This process gradually expands the receptive field while achieving a global statistical aggregation of local residual signals, thereby distilling highly discriminative high-frequency pattern features.

\subsubsection{Dynamic Feature Fusion Mechanism}
Upon obtaining the RGB feature vector $Z_r \in \mathbb{R}^d$ and the noise feature vector $Z_n \in \mathbb{R}^d$, the appropriate fusion of such heterogeneous representations also affects the ultimate discriminative efficacy of the model. To this end, we design a dynamic feature modulation mechanism guided by spatial features to achieve the deep fusion of the dual-branch representations.

As illustrated in Figure \ref{fig:architecture}, we adopt a modulation strategy based on FiLM. Using $Z_r$ as a priori input, a Multi-Layer Perceptron (MLP) dynamically generates modulation parameters. The MLP consists of two linear projection layers with channel dimensions expanding as $d \rightarrow 2d \rightarrow 3d$. It generates a scaling coefficient $\gamma$, a shift coefficient $\beta$, and a gating vector $g$, which can be formulated as
\begin{equation}
[\gamma, \beta, g] = f_{film}(Z_r) 
\end{equation}
where $f_{film}(\cdot)$ is the mapping that includes two linear transformations and a ReLU non-linear activation function. A Sigmoid function is applied to $g$ to ensure that the values of the gating weights fall within a reasonable and stable range. Subsequently, the generated affine parameters ($\gamma$, $\beta$) and the gating vector ($g$) are utilized to perform feature-wise linear modulation and secondary filtering on the noise feature $Z_n$:
\begin{equation}
\hat{Z}_n = g \odot (\gamma \odot Z_n + \beta)
\end{equation}
where $\odot$ denotes element-wise multiplication. In this integrated operation, the affine parameters $\gamma$ and $\beta$ perform precise recalibration and shifting of the low-level noise features driven by the conditional guidance, enabling the noise representation to adaptively align with the structural and content distribution of the current image. Concurrently, the gating vector $g$ acts as an information valve along the feature dimension to further suppress irrelevant noise interference. By dynamically controlling the transmission ratio of the modulated features, it prevents the excessive amplification of anomalous statistical artifacts. Ultimately, the joint cross-branch representation is constructed via a residual connection:
\begin{equation}
 Z_f = Z_r + \hat{Z}_n 
\end{equation}
Consequently, the final joint representation $Z_f \in \mathbb{R}^d$ accurately captures and amplifies intrinsic generative artifacts, providing a purer and more robust discriminative basis. To map these fused features into the final decision space, $Z_f$ is first processed by a sequence of dense layers, yielding a highly compact and abstract representation $Z_f'$. Ultimately, $Z_f'$ is fed into the classification head, which outputs the forgery probability via a Sigmoid activation function.

\subsubsection{Hard Sample-aware Contrastive Learning}
The conventional training paradigm typically relies on cross-entropy loss for global supervision. However, it overlooks the hard-to-distinguish samples near the decision boundary, thereby limiting the model's generalization capability against highly deceptive synthetic images. To attenuate such deficiency, we formulate a joint loss function comprising two components: a binary classification loss for branch supervision and a contrastive loss for constraining topological structure of the feature space. The overall objective function $\mathcal{L}_{total}$ is defined as
\begin{equation}
\mathcal{L}_{total} = \mathcal{L}_{cls} + \lambda \mathcal{L}_{contra}
\end{equation}
where $\mathcal{L}_{cls}$ is the classification loss, $\mathcal{L}_{contra}$ is the contrastive loss, and $\lambda$ is the weighting coefficient.

Following the protocol of supervised contrastive learning~\citep{khosla2020supervised}, for each anchor, samples with the same label form positive pairs, while those with different labels form negative pairs. During training, the gradient is dominated by easily distinguishable pairs, which causes the network to quickly hit an optimization bottleneck. The most critical "hard negatives"--samples that carry a different label yet lie perilously close to the anchor in the feature space--are virtually ignored. This leaves the model blind to the subtle, discriminative cues that separate authentic and generated images~\citep{RobinsonCSJ21}. To address this, we propose a Hard Sample-aware Contrastive Loss (HSCL) based on dynamic re-weighting. It adjusts the penalty adaptively for hard negative samples, guiding the network to focus its optimization on the most confusable boundary features.

The computation of HSCL proceeds as follows. Let $B$ be the total number of samples in the current mini-batch. For each sample $i$, the final embedding vector $z_i \in \mathbb{R}^{d}$ is first $L_2$-normalized to ensure that similarities are measured strictly by angular distance. Subsequently, the cosine similarity matrix $s$ between all pairs of samples within the batch is computed and scaled by a temperature parameter $\tau$ to control the smoothness of the distribution:
\begin{equation}
s_{i,j} = \frac{z_i \cdot z_j^T}{\tau}.
\end{equation}
Specifically, for each anchor $i$, $N(i)$ denotes the set of its negative samples. To counteract the dilution caused by easy negatives, we dynamically select the top-$K$ negatives in $N(i)$ with the highest similarity $s_{ij}$ and assign them an extra penalty. Denoting the exponentiated similarity as $\exp(s_{ij})$, the re-weighted negative term is defined as
\begin{equation}
\tilde{E}_{ij}^- = \exp(s_{ij}) \cdot \bigl(1 + \alpha H_{ij}\bigr)
\end{equation}
where $H_{ij} \in \{0,1\}$ indicates whether sample $j$ is among the selected hard negatives, and $\alpha$ controls the penalty intensity. For positive pairs, we simply set $E_{ij}^+ = \exp(s_{ij})$ when $i$ and $j$ share the same label (and $0$ otherwise). The proposed HSCL is then defined as
\begin{equation}
\mathcal{L}_{\text{contra}} = -\frac{1}{B} \sum_{i=1}^{B} \log \left( \frac{\sum_{j=1}^{B} E_{ij}^+}{\sum_{j=1}^{B} E_{ij}^+ + \sum_{j=1}^{B} \tilde{E}_{ij}^- + \epsilon} \right)
\end{equation}
where $\epsilon$ is a small constant for numerical stability. By amplifying the repulsion on those cross-class neighbors that lie nearest to the anchor, HSCL forces the feature boundaries to be sharpest in the most confusable regions. This yields a highly discriminative feature space and acts as an effective complement to the classification loss, significantly improving generalization ability to unseen generative models.

For the binary classification task, we employ the Binary Cross-Entropy (BCE) loss to measure the discrepancy between the predicted probabilities and the ground-truth labels. Assuming that for sample $i$, the model's predicted output is $p_i \in [0, 1]$ and the ground-truth label is $y_i \in \{0, 1\}$, the classification loss $\mathcal{L}_{cls}$ is defined as
\begin{equation}
\mathcal{L}_{cls} = -\frac{1}{B} \sum_{i=1}^{B} [y_i \cdot \log(p_i) + (1 - y_i) \cdot \log(1 - p_i)]
\end{equation}
Throughout the training process, the HSCL objective is responsible for structuring a well-clustered, highly separable feature space, while the BCE loss is tasked with delineating the optimal classification decision boundary within this optimized manifold.

\begin{figure}
    \centering
    \includegraphics[width=0.8\linewidth]{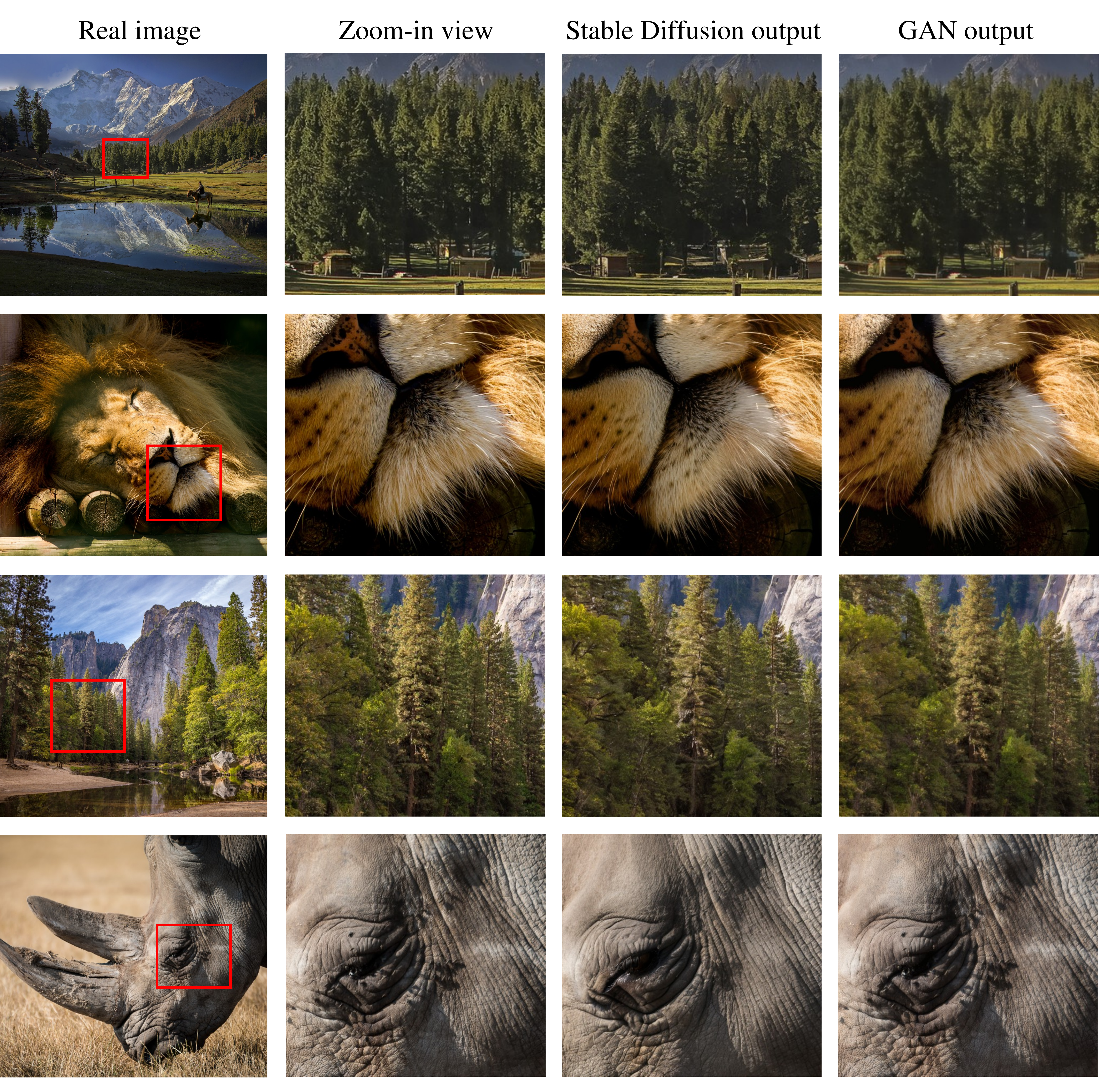} 
    \caption{Comparison of local details between synthetic and pristine images. From left to right: original image, zoomed-in local patch, SD XL~\cite{podell2024sdxl} generated image, and Real-ESRGAN~\cite{wang2021real} generated image.}
    \label{fig:detail_comparison}
\end{figure}

\subsection{Construction of Aligned Multi-source Synthetic Image Dataset (AMSID)}
Existing deepfake detection datasets often suffer from limited content diversity and outdated generative models, which cause models to overfit to dataset-specific biases rather than learning generalizable forgery artifacts. As observed by~\citet{cozzolino2024raising}, incorporating diverse real-image sources is crucial to bridging the generalization gap. To this end, we construct the Aligned Multi-source Synthetic Image Dataset (AMSID), a large-scale benchmark of approximately \textbf{235,000 images} designed to force the model to capture semantic-independent generation traces. By pairing diverse real images with their synthetic counterparts produced through modern generators, we explicitly eliminate content-driven distribution gaps and strengthen generalization.

AMSID collects pristine source images from six open-source datasets spanning a wide range of scenes, resolutions, and semantic domains: MS-COCO~\citep{lin2014microsoft}, MMP~\citep{chang2025mmp}, DIV2K~\citep{agustsson2017ntire}, BDD100K~\citep{yu2020bdd100k}, RAISE~\citep{dang2015raise}, and Flickr2K~\citep{lim2017enhanced}. To ensure that the network encounters a rich variety of generative fingerprints, we reconstruct these real images using two representative families of generators: Diffusion Models (DMs) and Generative Adversarial Networks (GANs). Specifically, Stable Diffusion XL (SD XL)~\citep{podell2024sdxl} is employed to regenerate images from the MS-COCO, MMP, and BDD100K subsets, injecting diffusion-specific artifacts. Real-ESRGAN~\citep{wang2021real} processes the DIV2K, RAISE, and Flickr2K subsets, imprinting GAN-related high-frequency texture biases. All synthetic images are generated strictly from their source counterparts, guaranteeing pixel-level alignment and identical semantic content. A visual comparison is provided in Figure~\ref{fig:detail_comparison}.

The generation pipeline is carefully controlled to preserve structural fidelity while maximizing the injection of model-intrinsic fingerprints. For the diffusion-based branch, we integrate the ControlNet Tile module~\citep{zhang2023adding} into the SD XL img2img pipeline to spatially confine local redrawing and prevent semantic drift. A DPM++ 2M sampler~\citep{lu2022dpm} with 30 steps and Karras scheduling~\citep{karras2022elucidating} is adopted for efficient high-quality synthesis, and the denoising strength is set to 0.25 to balance artifact visibility and global consistency. For the GAN-based branch, Real-ESRGAN’s high-order degradation modeling is leveraged in an end-to-end manner; it performs blind restoration at original resolutions, using its adversarial prior to re-infer and sharpen textures while naturally embedding the discriminator’s generative fingerprints. Both tracks produce synthetic images that are highly realistic and spatially aligned with the originals, yielding a clean, diverse training set explicitly built for learning generic forgery representations.

\section{Experiments}
\subsection{Experimental Settings}
\subsubsection{Datasets}
\label{sec:dataset}
\textbf{Training Datasets.} The training data configuration of our model is detailed in Table \ref{tab:training_set}. About \textbf{200,000 images} are collected from our created AMSID dataset and the self-conditioned subset of the public Bias-Free~\citep{guillaro2025bias} benchmark, which reconstructs MS-COCO~\citep{lin2014microsoft} images at pixel-level via Stable Diffusion 2.1~\citep{rombach2022high}. During the training phase, a strict paired alternating sampling is enforced to keep equal real and fake samples within each batch. Specifically, a pristine source image and its reconstructed synthetic counterpart are fed into the network in a mandated alternating sequence (i.e., $Real_A$, $Fake_A$, $Real_B$, $Fake_B$, \dots). This strictly paired formulation not only eliminates the distribution bias caused by varying semantic contents, but also provides sample pairs for contrastive learning. As such, the network could focus purely on capturing the microscopic generative artifacts.

\begin{table}[!b]
    \centering
    \caption{Detailed information of our training datasets. $^{*}$ means the typical spatial resolution of images in the dataset.}
    \label{tab:training_set}
    \vspace{0.1em}
    
    \setlength{\tabcolsep}{4pt} 
    \footnotesize
    
    \resizebox{\linewidth}{!}{
        \begin{tabular}{l l l | l | l | l}
            \toprule
            \textbf{Dataset} & \textbf{Generator} & \textbf{Source} & \textbf{Real / Fake} & \textbf{Resolution} & \textbf{Format} \\
            \midrule
            
            Bias-Free~\cite{guillaro2025bias}
            & SD 2.1~\cite{rombach2022high} & MS-COCO~\cite{lin2014microsoft} & 51,515 / 51,515 & 512 $\times$ 512 & PNG \\
            \midrule
            
            \multirow{6}{*}{AMSID}
            & \multirow{3}{*}{SD XL~\cite{podell2024sdxl}}
            & MS-COCO~\cite{lin2014microsoft}    & 27,935 / 27,935 & 480 $\times$ 640$^{*}$  & JPG / PNG \\
            &               & MMP~\cite{chang2025mmp}         & 2,000 / 2,000   & 768 $\times$ 1024$^{*}$ & JPG / PNG \\
            &               & BDD100K~\cite{yu2020bdd100k}    & 10,000 / 10,000 & 720 $\times$ 1280       & JPG / PNG \\
            \cmidrule(l){2-6}
            
            & \multirow{3}{*}{Real-ESRGAN~\cite{wang2021real}}
            & Flickr2K~\cite{lim2017enhanced}   & 2,650 / 2,650   & 1356 $\times$ 2040$^{*}$ & PNG \\
            &               & DIV2K~\cite{agustsson2017ntire} & 900 / 900       & 1356 $\times$ 2040$^{*}$ & PNG \\
            &               & RAISE~\cite{dang2015raise}      & 5,000 / 5,000   & 3264 $\times$ 4928$^{*}$ & TIF / PNG \\
            
            \bottomrule
        \end{tabular}
    }
\end{table}

\begin{table}[!t]
	\centering
	\caption{Summary of eight test datasets.}
	\label{tab:datasets}
	\vspace{0.1em}
	
	\footnotesize 
	\setlength{\tabcolsep}{4pt} 
	
	\resizebox{\linewidth}{!}{
		\begin{tabular}{@{}lllll@{}}
			\toprule
			\textbf{Datasets} & \textbf{Real/Fake} & \textbf{Source of Real}& \textbf{Generator Type}& \textbf{Generators}\\
			\midrule
			Synthbuster~\cite{bammey2023synthbuster} & 1K / 9K & RAISE & DM & 9 \\
			AIGCDetectionBenchmark~\cite{zhong2023patchcraft} & 74.3K / 74.3K & LSUN \& ImageNet, etc. & DM \& GAN & 16 \\
			UniversalFakeDetect~\cite{ojha2023towards} & 52K / 52K & LAION \& ImageNet, etc. & DM \& GAN & 18 \\
			GenImage~\cite{zhu2023genimage} & 50K / 50K & ImageNet & DM \& GAN & 8 \\
			DDA-COCO~\cite{chen2025dual} & 5K / 25K & MSCOCO & DM & 5 \\
			DIF~\cite{sinitsa2023deep} & 37.7K / 37.7K & LAION & DM \& GAN & 13 \\
			Chameleon~\cite{yan2025a} & 14.9K / 11.2K & Internet & Unknown & Unknown \\
			WildRF~\cite{cavia2024real} & 1.1K / 1.2K & Reddit, FB, X & Unknown & Unknown \\
			\bottomrule
		\end{tabular}
	}
\end{table}

\textbf{Data Augmentation.} To enhance the robustness against real-world degradations and prevent overfitting to superficial artifacts, we employ a probability-driven data augmentation pipeline comprising four sequential operations:
\begin{enumerate}[label=\arabic*)]
    \item \textbf{Center Cropping:} Input images are center-cropped to $224 \times 224$ pixels to standardize dimensions and ensure semantic alignment.
    \item \textbf{Geometric Transformation:} Applied with a probability of $0.5$, randomly selecting a mutually exclusive operation (horizontal/vertical flip, $90^{\circ}$ rotation, or transposition) to foster spatial invariance.
    \item \textbf{Image Compression:} JPEG compression (QF $\in [30, 100]$) is applied with a probability of $0.5$. We randomly alternate between OpenCV and PIL encoders to emulate diverse cross-platform transmission artifacts.
    \item \textbf{Gaussian Blurring:} Gaussian blur ($\sigma \in [0.0, 3.0]$) is applied with a probability of $0.5$ to obscure fragile high-frequency noise, compelling the network to capture robust, degradation-invariant structural representations.
\end{enumerate}

\textbf{Testing Datasets.} To comprehensively evaluate the cross-domain generalization of our model, we employ 8 public benchmark datasets, i.e., UniversalFakeDetect~\citep{ojha2023towards}, AIGCDetectionBenchmark~\citep{zhong2023patchcraft}, Synthbuster~\citep{bammey2023synthbuster}, GenImage~\citep{zhu2023genimage}, DIF~\citep{sinitsa2023deep}, DDA-COCO~\citep{chen2025dual}, Chameleon~\citep{yan2025a} and WildRF~\citep{cavia2024real}. The detailed compositions of these datasets are summarized in Table~\ref{tab:datasets}. The final test set contains approximately \textbf{500,000 images}, spanning Diffusion Models, GANs, and unknown commercial generators.

\subsubsection{Baseline Methods}
Our proposed method is compared with $8$ representative state-of-the-art baselines across diverse detection paradigms, which include the gradient-based~\citep{tan2023learning}, reconstruction-based~\citep{wang2023dire,chen2024drct}, semantic-based~\citep{ojha2023towards,tan2025c2p,cozzolino2024raising}, and multi-domain fusion-based~\citep{yan2025a} approaches. These methods are briefly introduced as follows:

\begin{itemize}
    \item \textbf{LGrad}~\citep{tan2023learning} (CVPR’2023) transforms input images into gradient maps via a pre-trained CNN to explicitly isolate underlying generative artifacts from high-level semantic content.
    
    \item \textbf{DIRE}~\citep{wang2023dire} (ICCV'2023) utilizes the discrepancy between the original image and its reconstructed counterpart produced by a pre-trained diffusion model as a discriminative metric.
    
    \item \textbf{UFD}~\citep{ojha2023towards} (CVPR’2023) leverages the robust and generalized representation space of a pre-trained CLIP model, combined with a K-Nearest Neighbors (KNN) classifier, to achieve cross-model detection.
    
    \item \textbf{Cozzo2024}~\citep{cozzolino2024raising} (CVPR’2024) constructs semantic-aligned pairs of real/generated images and performs similarity discrimination in the CLIP feature space to achieve effective few-shot forgery detection.
    
    \item \textbf{RINE}~\citep{koutlis2024leveraging} (ECCV'2024) extracts multi-level representations from the intermediate Transformer blocks of the CLIP encoder to construct a highly discriminative forgery-aware feature space.
    
    \item \textbf{DRCT}~\citep{chen2024drct} (ICML'2024) applies contrastive learning on both the original images and their diffusion-reconstructed trajectories to explicitly widen the representation margin between the pristine and synthetic domains.

    \item \textbf{AIDE}~\citep{yan2025a} (ICLR'2025) constructs a hybrid dual-branch detector by dynamically fusing global semantic features extracted by CLIP with local spatial representations derived from DCT coefficients.

    \item \textbf{C2P-CLIP}~\citep{tan2025c2p} (AAAI'2025) injects category-common prompts into the CLIP architecture and fine-tunes the model via contrastive learning to strengthen the real-vs-fake decision boundary.
\end{itemize}

\subsubsection{Implementation Details}
The proposed RNSIDNet is implemented using PyTorch and trained on a single NVIDIA RTX 3090 GPU. The specific data augmentation and pre-processing strategies for generating the $224 \times 224$ image inputs strictly follow the protocols detailed in Section \ref{sec:dataset}. In the dual-branch architecture, the RGB branch employs a frozen CLIP ViT-L/14 backbone to extract global semantics, and the noise branch utilizes $5 \times 5$ Bayar convolutional kernels to capture high-frequency artifacts. Prior to dynamic fusion, the extracted representations from both branches are mapped and aligned into a unified $d=512$ dimensional space.

The network is trained for 1 epochs using the standard Adam optimizer with a batch size of 64. The learning rate is initialized at $1 \times 10^{-4}$. Crucially, to implement our HSCL strategy, the balancing coefficient $\lambda$ for the joint objective is empirically set to $0.8$, and the contrastive temperature $\tau$ is set to $0.07$. To effectively penalize highly deceptive samples, the top $10\%$ of negative samples within each batch are dynamically mined as "hard negatives" and assigned an explicit penalty weight of $2.0$.

\subsubsection{Evaluation Metrics}
Following standard image forensics protocols, we employ Accuracy (ACC) and the Area Under the Receiver Operating Characteristic Curve (AUC) as our primary quantitative metrics. The classification threshold is fixed at $0.5$. ACC measures the overall classification correctness, while AUC provides a robust evaluation of the model's discriminative capability across varying thresholds by integrating the True Positive Rate (TPR) against the False Positive Rate (FPR). Meanwhile, We also report the average (AVG) metric values across the test datasets to obtain summary evaluations.

\begin{table}[!t]
	\centering
	\caption{Performance comparison of various models across all evaluation benchmarks. Results are presented in the format of ACC (\%) / AUC (\%). Due to table width constraints, AIGCDetectionBenchmark~\cite{zhong2023patchcraft} and UniversalFakeDetect~\cite{ojha2023towards} are abbreviated as AIGCDetect and UFDetect, respectively. The best and second-best results are highlighted in bold and underlined.}
	\label{tab:rnsidnet_performance}
	\vspace{0.1em}
	
	\scriptsize 
	\setlength{\tabcolsep}{3.5pt} 
	
	\resizebox{\linewidth}{!}{
		\begin{tabular}{@{}lccccccccc@{}}
			\toprule
			Method & GenImage & Synthbuster & AIGCDetect & UFDetect & DDA-COCO & DIF & Chameleon & WildRF & AVG \\
			\midrule
			LGrad~\cite{tan2023learning} & 51.94/57.04 & 44.33/39.16 & 50.19/53.81 & 36.51/35.94 & 50.19/52.15 & 49.91/53.88 & 48.49/47.43 & 47.79/60.67 & 47.42/50.01 \\
			
			DIRE~\cite{wang2023dire} & 56.05/61.83 & 47.80/46.88 & 53.74/56.33 & 49.98/48.61 & 51.41/52.10 & 52.01/53.46 & 46.66/51.79 & 53.73/54.26 & 51.42/53.16 \\
			
			UFD~\cite{ojha2023towards} & 69.87/88.67 & 55.36/53.94 & 78.43/91.75 & 82.93/93.42 & 52.75/71.15 & 81.44/91.11 & 57.36/54.29 & 55.60/58.92 & 66.72/75.41 \\
			
			DRCT~\cite{chen2024drct} & 83.46/98.08 & 72.99/74.42 & 70.67/80.05 & 70.90/79.12 & \underline{70.23}/85.59 & 68.49/76.94 & \textbf{69.90}/\underline{74.48}& 68.14/74.05 & 71.85/80.34 \\
			
			Cozzo2024~\cite{cozzolino2024raising} & 81.79/94.81 & 73.86/\underline{91.60} & 77.49/90.33 & 67.88/80.98 & 50.42/61.25 & 75.93/86.15 & 55.37/52.83 & \textbf{73.98}/\textbf{89.55} & 69.59/80.94 \\
			
			RINE~\cite{koutlis2024leveraging} & \underline{95.15}/\underline{99.25} & \underline{87.49}/87.54 & \underline{90.10}/\textbf{98.58} & \textbf{88.34}/\textbf{97.88} & 51.36/\underline{85.96} & \textbf{89.54}/\textbf{98.21} & 46.97/45.43 & 72.20/79.64 & \underline{77.64}/\underline{86.56} \\
			
			AIDE~\cite{yan2025a} & 87.14/96.78 & 56.26/62.49 & 81.94/92.40 & 78.65/88.32 & 50.12/54.74 & 80.08/90.65 & 64.20/73.60& 66.04/73.71 & 70.55/79.09 \\
			
			C2P-CLIP~\cite{tan2025c2p} & \textbf{96.59}/\textbf{99.51} & 46.12/54.50 & 83.47/90.81 & 74.29/86.06 & 50.95/66.92 & 75.38/86.00 & 54.14/59.77 & 59.57/67.23 & 67.56/76.35 \\
			
			\midrule
			
			RNSIDNet & 91.70/98.40& \textbf{92.43}/\textbf{97.38} & \textbf{91.32}/\underline{97.61}& \underline{83.94}/\underline{94.13}& \textbf{85.18}/\textbf{98.22}& \underline{86.18}/\underline{95.78}& \underline{66.65}/\textbf{76.80}& \underline{73.85}/\underline{83.46}& \textbf{83.81}/\textbf{92.72}\\
			\bottomrule
	\end{tabular}}
\end{table}

\begin{table}[!t]
	\centering
	\caption{Performance comparison of different AI-generated image detection models on the AIGCDetectionBenchmark~\citep{zhong2023patchcraft} dataset. All benchmark results are reported as ACC (\%).}
	\label{tab:performance_comparison_acc}
	
	\vspace{0.1em}
	\scriptsize
	\setlength{\tabcolsep}{1.5pt}
	
	\resizebox{\linewidth}{!}{
		\begin{tabular}{lccccccccccccccccc}
			\toprule
			Method & ADM & DALL-E 2 & Glide & Midjourney & VQDM & BigGAN & CycleGAN & GauGAN & ProGAN & SD 1.4 & SD 1.5 & StarGAN & StyleGAN & StyleGAN 2 & WFR & WuKong & AVG \\
			\midrule
			LGrad~\cite{tan2023learning} & 53.45 & 57.60 & 60.05 & 56.83 & 51.40 & 48.18 & 44.17 & 48.10 & 49.20 & 47.05 & 46.75 & 48.17 & 48.15 & 46.04 & 48.55 & 49.30 & 50.19 \\
			
			DIRE~\cite{wang2023dire} & 57.63 & 69.45 & 62.92 & 58.00 & 54.91 & 46.42 & 50.11 & 52.26 & 49.98 & 51.35 & 51.82 & 50.15 & 50.29 & 52.96 & 49.50 & 52.06 & 53.74 \\
			
			UFD~\cite{ojha2023towards} & 66.87 & 50.75 & 62.46 & 56.13 & 85.31 & \underline{95.08} & \textbf{98.33} & \textbf{99.47} & \textbf{99.81} & 63.66 & 63.49 & \underline{95.75} & 84.93 & 74.96 & \underline{86.90} & 70.93 & 78.43 \\
			
			DRCT~\cite{chen2024drct} & 66.38 & 77.55 & 73.24 & \underline{94.33} & 76.78 & 60.15 & 49.77 & 50.94 & 58.53 & \underline{99.24} & \underline{99.11} & 55.48 & 64.40 & 55.05 & 50.50 & \textbf{99.22} & 70.67 \\
			
			Cozzo2024~\cite{cozzolino2024raising} & 66.58 & 90.75 & \underline{95.93} & 68.08 & 82.75 & 74.10 & 87.28 & 83.94 & 72.96 & 85.46 & 85.72 & 54.95 & 70.50 & 70.86 & 71.55 & 78.50 & 77.49 \\
			
			RINE~\cite{koutlis2024leveraging} & \textbf{95.65} & 74.80 & 92.07 & 81.67 & \textbf{96.58} & 88.88 & 94.40 & \underline{98.09} & \underline{99.49} & 96.76 & 96.62 & 63.33 & \underline{87.39} & \underline{79.51} & \textbf{99.60} & 96.68 & \underline{90.10} \\
			
			AIDE~\cite{yan2025a} & 78.47 & \underline{95.00} & 91.76 & 81.42 & 80.25 & 77.20 & 74.38 & 64.36 & 69.27 & \textbf{99.74} & \textbf{99.74} & 80.27 & 71.32 & 72.51 & 76.45 & \underline{98.85} & 81.94 \\
            
			C2P-CLIP~\cite{tan2025c2p} & \underline{90.22} & \textbf{99.35} & \textbf{97.81} & \textbf{97.55} & \underline{96.39} & 72.12 & \underline{97.09} & 68.39 & 56.49 & 98.99 & 98.79 & 61.31 & 61.54 & 58.33 & 82.45 & 98.62 & 83.47 \\
			
			\midrule
			
			RNSIDNet & 83.73 & 87.30 & 89.93 & 94.32 & 93.14 & \textbf{97.25} & 96.03 & 97.80 & 90.29 & 96.92 & 96.96 &  \underline{95.02} & \underline{86.93} & \textbf{86.52} & 74.20 & 94.79 & \textbf{91.32} \\
			\bottomrule
	\end{tabular}}
\end{table}

\begin{table}[!t]
	\centering
	\caption{Performance comparison of different AI-generated image detection models on the AIGCDetectionBenchmark~\cite{zhong2023patchcraft} dataset. All benchmark results are reported as AUC (\%).}
	\label{tab:performance_comparison_auc}
	
	\vspace{0.1em}
	\scriptsize
	\setlength{\tabcolsep}{1.5pt}
	
	\resizebox{\linewidth}{!}{
		\begin{tabular}{lccccccccccccccccc}
			\toprule
			Method & ADM & DALL-E 2 & Glide & Midjourney & VQDM & BigGAN & CycleGAN & GauGAN & ProGAN & SD 1.4 & SD 1.5 & StarGAN & StyleGAN & StyleGAN 2 & WFR & WuKong & AVG \\
			\midrule
			LGrad~\cite{tan2023learning} & 60.80 & 77.11 & 71.46 & 64.44 & 52.22 & 46.10 & 48.37 & 52.91 & 50.81 & 47.78 & 47.68 & 46.58 & 48.21 & 45.09 & 52.08 & 49.36 & 53.81 \\
			
			DIRE~\cite{wang2023dire} & 63.35 & 82.25 & 78.66 & 64.12 & 59.61 & 46.00 & 47.34 & 52.27 & 49.64 & 53.02 & 53.64 & 45.15 & 51.78 & 52.00 & 48.16 & 54.36 & 56.33 \\
			
			UFD~\cite{ojha2023towards} & 87.20 & 66.82 & 85.24 & 75.54 & 96.49 & 99.19 & \underline{99.77} & \textbf{99.98} & \textbf{100.00} & 87.79 & 87.35 & 99.32 & \underline{97.29} & \underline{97.93} & \underline{96.52} & 91.59 & 91.75 \\
			
			DRCT~\cite{chen2024drct} & 94.84 & 99.69 & 97.52 & \underline{99.33} & 96.94 & 70.65 & 62.27 & 47.58 & 63.89 & \textbf{100.00} & \underline{99.98} & 59.91 & 74.32 & 59.96 & 53.92 & \textbf{99.99} & 80.05 \\
			
			Cozzo2024~\cite{cozzolino2024raising} & 85.85 & \underline{99.85} & \underline{99.35} & 89.17 & 96.22 & 81.89 & 92.41 & 92.07 & 92.58 & 97.22 & 97.30 & 81.22 & 78.43 & 77.65 & 89.06 & 94.94 & 90.33 \\
			
			RINE~\cite{koutlis2024leveraging} & \textbf{99.25} & 99.34 & 97.68 & 95.01 & \textbf{99.75} & \textbf{99.81} & \underline{99.91} & \textbf{99.98} & \textbf{100.00} & 99.89 & 99.85 & \underline{99.98} & \textbf{98.45} & 88.52 & \textbf{99.99} & 99.83 & \textbf{98.58} \\
			
			AIDE~\cite{yan2025a} & 91.46 & \textbf{99.92} & 98.54 & 97.40 & 96.00 & 85.67 & 94.55 & 72.57 & 84.25 & \underline{99.98} & \textbf{99.99} & 91.07 & 87.22 & 90.37 & 89.38 & \underline{99.96} & 92.40 \\
            
			C2P-CLIP~\cite{tan2025c2p} & \underline{98.58} & \underline{99.85} & \textbf{99.50} & \textbf{99.68} & \underline{99.40} & 83.65 & 98.91 & 76.48 & 83.22 & 99.97 & 99.96 & 89.07 & 68.03 & 65.87 & 90.88 & 99.83 & 90.81 \\
			\midrule
			
			RNSIDNet & 96.03 & 99.22 & 97.85 & 99.18 & 98.83 & \underline{99.71} & 99.36 & \underline{99.89} & \underline{99.79} & 99.67 & 99.59 & \textbf{99.99} & 96.47 & \underline{93.73} & 83.30 & 99.18 & \underline{97.61} \\
			\bottomrule
	\end{tabular}}
\end{table}

\subsection{Comparison to State-of-the-art Models}
Extensive evaluations across eight benchmarks (Tables~\ref{tab:rnsidnet_performance}--\ref{tab:performance_ddacoco}) demonstrate that RNSIDNet achieves state-of-the-art cross-domain generalization, with an average ACC of 83.81\% and AUC of 92.72\%. To expose common failure modes, we closely examine several representative test sets. AIGCDetectionBenchmark~\citep{zhong2023patchcraft} is the largest benchmark dataset in our experiment set with the widest variety of generators, while Synthbuster~\citep{bammey2023synthbuster} features high-resolution images synthesized by recent advanced generative models. In our experiments, several strong baselines suffer a catastrophic performance inversion. For instance, the detection ACC of C2P-CLIP~\citep{tan2025c2p} falls sharply from 83.47\% to 46.12\%. Such drastic variance signals that RGB-only or prior-dependent detectors tend to memorize generator-specific superficial cues, leading to severe modality overfitting and a form of representation collapse under distribution shift. In contrast, our method remains remarkably stable across this spectrum, effectively bypassing the semantic camouflage that deceives single-modality detectors.

\begin{table}[!t]
	\centering
	\caption{Performance comparison of different AI-generated image detection models on the Synthbuster~\cite{bammey2023synthbuster} dataset. All benchmark results are reported as ACC (\%) / AUC (\%).}
	\label{tab:performance_synthbuster}
	\vspace{0.1em}
	
	\scriptsize 
	\setlength{\tabcolsep}{2pt}
	
	\resizebox{\linewidth}{!}{
		\begin{tabular}{@{}lcccccccccc@{}}
			\toprule
			Method & Dalle-E 2 & Dalle-E 3 & Firefly & Glide & Midjourney & SD 1.3 & SD 1.4 & SD 2 & SD XL & AVG \\
			\midrule
			LGrad~\cite{tan2023learning} & 48.15/47.71 & 40.80/37.72 & 41.85/37.84 & 61.45/64.41 & 44.95/47.89 & 40.30/29.08 & 40.25/28.47 & 40.70/23.53 & 40.50/35.82 & 44.33/39.16 \\
			
			DIRE~\cite{wang2023dire} & 53.55/56.29 & 36.35/32.14 & 51.45/48.68 & 60.90/69.87 & 51.90/51.26 & 35.05/29.76 & 34.60/29.38 & 50.10/48.30 & 56.30/56.27 & 47.80/46.88 \\
			
			UFD~\cite{ojha2023towards} & 70.95/78.59 & 34.80/9.02 & 78.15/88.40 & 40.70/34.40 & 39.90/28.85 & 57.85/60.71 & 57.35/60.65 & 63.10/67.71 & 55.45/57.12 & 55.36/53.94 \\
			
			DRCT~\cite{chen2024drct} & 47.55/36.28 & 53.40/46.49 & 48.10/35.37 & 60.20/73.86 & 89.30/95.64 & 94.10/\textbf{100.00} & 94.10/\textbf{100.00} & 90.80/96.44 & 79.40/85.69 & 72.99/74.42 \\
			
			Cozzo2024~\cite{cozzolino2024raising} & 68.10/87.49 & \underline{74.15}/\textbf{93.69} & 64.55/88.38 & \textbf{97.25}/\textbf{99.65} & 62.85/83.18 & 78.85/94.87 & 78.55/94.66 & 71.45/91.35 & 68.95/91.14 & 73.86/\underline{91.60} \\
			
			RINE~\cite{koutlis2024leveraging} & \underline{89.80}/\underline{95.28} & 47.20/11.36 & \textbf{85.25}/\textbf{92.22} & \underline{90.05}/\underline{94.89} & \underline{92.45}/\underline{96.32} & \underline{96.45}/\textbf{100.00} & \textbf{96.45}/\textbf{100.00} & \underline{93.50}/\underline{98.08} & \underline{96.30}/\underline{99.74} & \underline{87.49}/87.54 \\
			
			AIDE~\cite{yan2025a} & 39.25/42.21 & 39.00/41.68 & 25.75/10.52 & 65.90/74.59 & 59.70/66.19 & 75.15/92.18 & 74.90/90.91 & 55.95/62.25 & 70.70/81.91 & 56.26/62.49 \\
			
			C2P-CLIP~\cite{tan2025c2p} & 49.60/57.58 & 49.85/65.75 & 17.25/2.45 & 49.90/36.23 & 49.90/62.10 & 50.10/79.30 & 50.10/78.99 & 48.35/51.74 & 50.00/56.38 & 46.12/54.50 \\
			
			\midrule
			
			RNSIDNet & \textbf{92.30}/\textbf{98.20} & \textbf{80.10}/\underline{90.66} & \underline{94.40}/\underline{98.79} & 81.10/89.95 & \textbf{97.15}/\textbf{99.90} & \textbf{96.50}/\underline{99.66} & \underline{95.90}/\underline{99.39} & \textbf{97.20}/\textbf{99.91} & \textbf{97.20}/\textbf{99.94} & \textbf{92.43}/\textbf{97.38} \\
			\bottomrule
	\end{tabular}}
\end{table}

\begin{table}[!t]
	\centering
	\caption{Performance comparison of different AI-generated image detection models on the DDA-COCO~\cite{chen2025dual} dataset. All benchmark results are reported as ACC (\%) / AUC (\%).}
	\label{tab:performance_ddacoco}
	\vspace{0.1em}
	
	\small
	\setlength{\tabcolsep}{14pt} 
	
	\resizebox{\linewidth}{!}{
		\begin{tabular}{@{}lcccccc@{}} 
		\toprule
		Method & SD-VAE-FT-EMA & SD-VAE-FT-MSE & SD XL-VAE & SD 2.1 & SD 3.5-Large & AVG \\
		\midrule
		LGrad~\cite{tan2023learning} & 50.22/51.08 & 50.63/54.88 & 50.00/51.17 & 50.63/54.87 & 49.47/48.73 & 50.19/52.15 \\
		
		DIRE~\cite{wang2023dire} & 50.32/50.47 & 52.38/53.55 & 51.95/53.14 & 52.40/53.55 & 50.00/49.79 & 51.41/52.10 \\
		
		UFD~\cite{ojha2023towards} & 54.40/76.50 & 53.24/73.59 & 51.52/66.35 & 53.29/73.59 & 51.29/65.72 & 52.75/71.15 \\
		
		DRCT~\cite{chen2024drct} & \underline{83.32}/\underline{95.27} & \underline{77.00}/\underline{93.11} & \underline{62.49}/\underline{81.69} & \underline{76.83}/\underline{93.10} & \underline{51.49}/64.76 & \underline{70.23}/85.59 \\
		
		Cozzo2024~\cite{cozzolino2024raising} & 50.46/63.99 & 50.55/62.97 & 50.32/60.92 & 50.57/62.93 & 50.18/55.43 & 50.42/61.25 \\
		
		RINE~\cite{koutlis2024leveraging} & 51.56/88.42 & 51.91/88.98 & 50.24/80.09 & 52.02/88.98 & 51.05/\underline{83.31} & 51.36/\underline{85.96} \\
		
		AIDE~\cite{yan2025a} & 50.18/56.83 & 50.14/55.95 & 50.13/53.10 & 50.12/55.95 & 50.05/51.88 & 50.12/54.74 \\
		
		C2P-CLIP~\cite{tan2025c2p} & 51.85/69.16 & 51.06/67.68 & 49.85/62.38 & 51.00/67.66 & 50.97/67.70 & 50.95/66.92 \\
		
		\midrule
		
		RNSIDNet & \textbf{84.18}/\textbf{97.95} & \textbf{89.69}/\textbf{98.91} & \textbf{86.05}/\textbf{98.48} & \textbf{89.63}/\textbf{98.91} & \textbf{76.35}/\textbf{96.84} & \textbf{85.18}/\textbf{98.22} \\
		\bottomrule
	\end{tabular}}
\end{table}

To further test generalization performance against cutting-edge generative architectures and extreme visual disguise, we evaluate on the DDA-COCO~\citep{chen2025dual} dataset. With pixel-level aligned fake images almost indistinguishable from real ones, prior-dependent methods like AIDE~\cite{yan2025a} degenerate to random guessing, while RNSIDNet maintains robust discrimination (80.94\% ACC, 98.00\% AUC). Notably, DRCT~\cite{chen2024drct} and RINE~\cite{koutlis2024leveraging} also exhibit competitive performance here. Given that both their and our method employ contrastive learning, this shared success underscores the critical role of contrastive constraints in disentangling subtle generation artifacts. Finally, in uncontrolled wild scenarios, RNSIDNet achieves the highest AUC of 76.80\% on Chameleon~\citep{yan2025a}, with the second-best ACC and AUC of on WildRF~\citep{cavia2024real}, confirming that the intrinsic physical fingerprints captured by our enhanced architecture are highly robust against complex real-world post-processing, bridging the gap between laboratory tests and open-world deployment.

\subsection{Ablation Studies}
\label{sec:ablation}

To rigorously validate the individual contributions of the core components in RNSIDNet, we conduct a comprehensive ablation study. This includes both leave-one-out experiments, which systematically remove the BAM, FiLM, or HSCL modules, and replacement experiments, which substitute the noise extractor and the contrastive loss function.

\begin{table*}[t]
\centering
\caption{Ablation study of RNSIDNet framework on the Synthbuster~\cite{bammey2023synthbuster} dataset. `w/o' denotes removing a specific module, while `w/' denotes replacing our proposed module with a conventional alternative.}
\label{tab:ablation}
\renewcommand{\arraystretch}{1.2} 
\setlength{\tabcolsep}{12pt}      
\begin{tabular}{l l c c c c}
\toprule
\textbf{Category} & \textbf{Model Variant} & \textbf{ACC (\%)} & $\Delta$ \textbf{ACC} & \textbf{AUC (\%)} & $\Delta$ \textbf{AUC} \\
\midrule
\textbf{Full Model} & \textbf{RNSIDNet (Ours)} & \textbf{92.92} & -- & 97.68 & -- \\
\midrule
\multirow{3}{*}{\textit{Module Ablation}} & w/o BAM & 90.33 & -2.59 & 97.21 & -0.47 \\
 & w/o FiLM  & 89.79 & -3.13 & 95.51 & -2.17 \\
 & w/o HSCL  & 90.42 & -2.50 & 96.46 & -1.22 \\
\midrule
\multirow{2}{*}{\textit{Noise Extractor}} & w/ SRM Filters & 91.81 & -1.11 & 97.43 & -0.25 \\
 & w/ Noiseprint++ & 88.97 & -3.95 & \textbf{97.96} & +0.28 \\
\midrule
\multirow{2}{*}{\textit{Loss Function}} & w/ InfoNCE++ & 90.41 & -2.51 & 96.58 & -1.10 \\
 & w/ SupCon & 90.62 & -2.30 & 96.72 & -0.96 \\
\bottomrule
\end{tabular}
\end{table*}

As shown in Table \ref{tab:ablation}, removing BAM, FiLM, or HSCL decreases the overall accuracy (92.92\%) to 90.33\%, 89.79\%, and 90.42\%, respectively. The most severe decline occurs without FiLM (a 3.13\% drop), underscoring its critical role in dynamically fusing heterogeneous modalities. Meanwhile, the performance degradation without BAM and HSCL confirms their necessity in refining semantic features and shaping precise decision boundaries. For the replacement variants, substituting Bayar convolutions with fixed SRM~\cite{fridrich2012rich} or Noiseprint++~\cite{Guillaro2023trufor} reduces the ACC to 91.81\% and 88.97\%. Although Noiseprint++ yields a marginally higher AUC (97.96\%), its inferior ACC suggests threshold instability, demonstrating that our learnable Bayar constraints are inherently more robust. Similarly, replacing HSCL with InfoNCE++~\cite{oord2018representation} or SupCon~\cite{khosla2020supervised} leads to sub-optimal performance, highlighting the advantage of our hard sample-aware mechanism over standard contrastive objectives.

\begin{figure}
    \centering
    \includegraphics[width=1\linewidth]{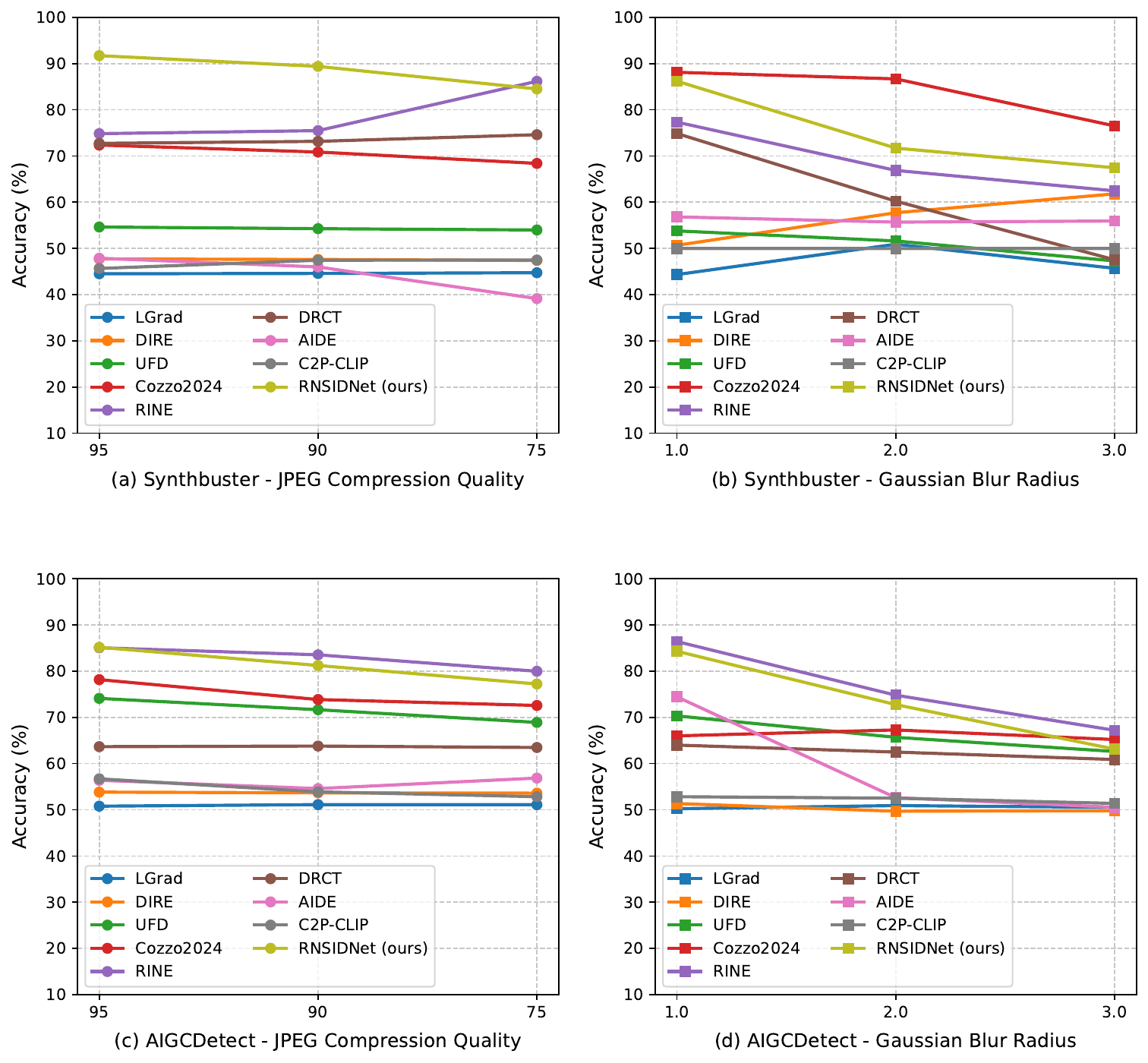} 
    \caption{Performance evaluation of different models under JPEG compression and Gaussian blur on Synthbuster~\cite{bammey2023synthbuster} and AIGCDetectionBenchmark~\citep{zhong2023patchcraft} datasets.}
    \label{fig:model_robustness_evaluation}
\end{figure}

\subsection{Robustness Analysis}
To verify the structural stability of our enhanced RGB-Noise representations, we conduct a comprehensive robustness analysis under various degradation conditions. Specifically, we evaluate the models on two distinct datasets, Synthbuster~\cite{bammey2023synthbuster} and AIGCDetect~\citep{zhong2023patchcraft} , applying JPEG compression and Gaussian blur across varying levels of intensity.

As illustrated in Figure~\ref{fig:model_robustness_evaluation}, RNSIDNet consistently maintains a competitive advantage across various degradation conditions and data sources. In contrast, several baseline methods exhibit significant instability and severe performance fluctuations due to different datasets. Meanwhile, we observe that the model inevitably experiences a performance drop under Gaussian blur perturbations. This is expected, as Gaussian blur acts as a low-pass filter that severely erases critical high-frequency information in the frequency domain. Nevertheless, the dual-branch architecture of RNSIDNet provides essential structural resilience, successfully preventing a complete performance collapse and sustaining a functional detection capability even under heavy blur.

\subsection{Scalability and Complexity Analysis}
\begin{figure}
	\centering
	\includegraphics[width=0.95\linewidth, height=0.4\textheight,keepaspectratio]{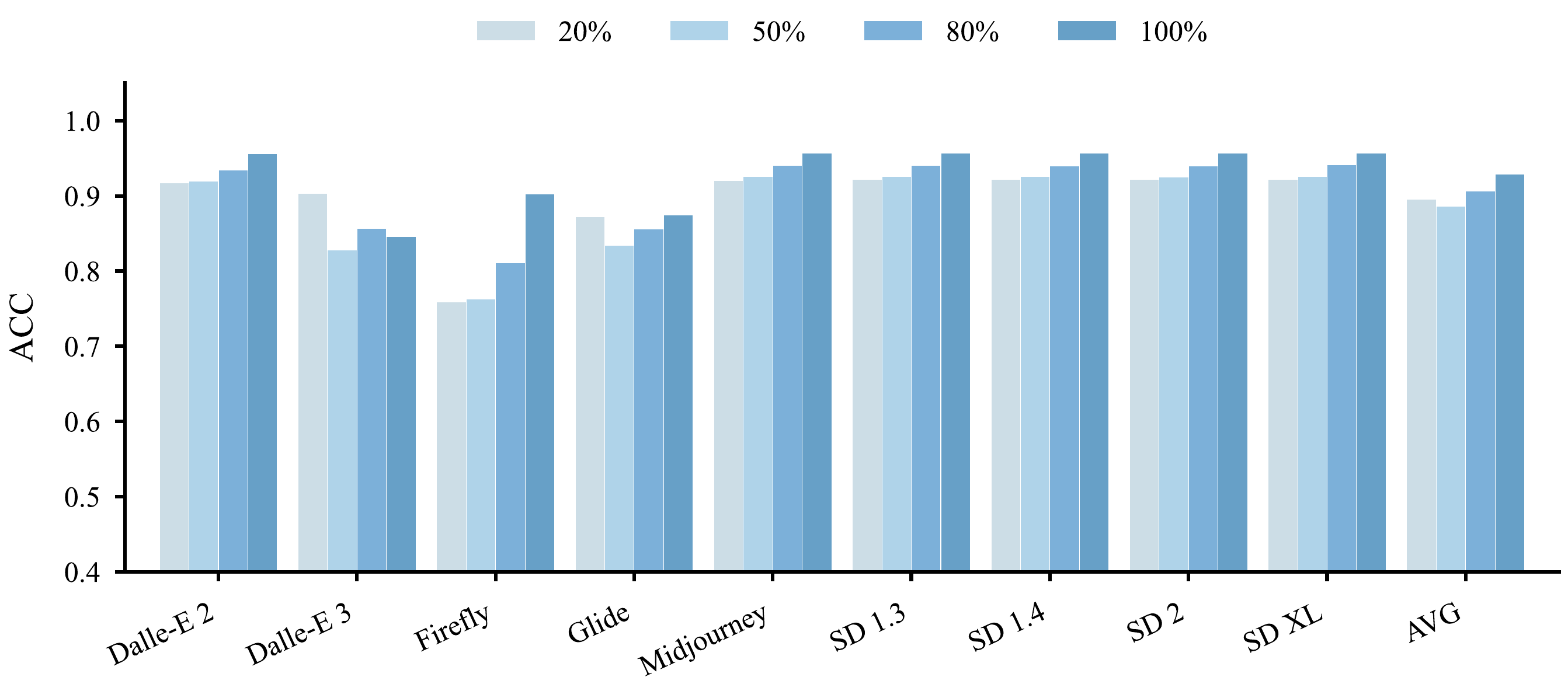} 
	\caption{Performance comparison of the RNSIDNet model trained with different sample set sizes on the Synthbuster~\cite{bammey2023synthbuster} dataset.}
	\label{fig:data_acc}
\end{figure}

\begin{table}[!t]
	\centering
	\caption{Comparison of model complexity among different methods.}
	\label{tab:complexity}
	\small
	\begin{tabular}{lccc}
		\toprule
		Method & Trainable Parameters (M) & Total Parameters (M) & FLOPs (G) \\
		\midrule
		LGrad~\cite{tan2023learning} & 23.51 & 23.51 & 8.26 \\
		DIRE~\cite{wang2023dire} & 23.51 & 23.51 & 8.26 \\
		UFD~\cite{ojha2023towards} & 0.00077 & 427.62 & 103.79 \\
		DRCT~\cite{chen2024drct} & 88.62 & 88.62 & 30.71 \\
		Cozzo2024~\cite{cozzolino2024raising} & 0.00077 & 427.62 & 103.79 \\
		RINE~\cite{koutlis2024leveraging} & 10.52 & 438.14 & 104.00 \\
		AIDE~\cite{yan2025a} & 54.43 & 897.83 & 451.39 \\
		C2P-CLIP~\cite{tan2025c2p} & 2.36 & 429.98 & 155.64 \\
		\midrule
		RNSIDNet & 13.58 & 441.20 & 104.31 \\
		\bottomrule
	\end{tabular}
\end{table}

In this section, We evaluate the practical applicability of RNSIDNet from two perspectives: data scalability and computational complexity. To assess data scalability, we trained the model using 20\%, 50\%, 80\%, and 100\% of the training data and evaluated it on the Synthbuster~\citep{bammey2023synthbuster} dataset. As illustrated in Figure~\ref{fig:data_acc}, increasing the training data boosts the average accuracy from 89.54\% to 92.92\%, enabling the network to learn more robust features. Notably, the model demonstrates excellent data efficiency, achieving over 90\% accuracy on Midjourney and Stable Diffusion with only 20\% of the data. However, handling more complex distributions requires more data; for instance, performance on Firefly surges from 75.90\% to 90.25\% when utilizing the full dataset. Thus, expanding the training set effectively enhances both absolute performance and cross-domain generalization.

Beyond data efficiency, Table~\ref{tab:complexity} compares the computational complexity of various methods. RNSIDNet achieves an optimal balance between efficiency and detection accuracy. While lightweight methods (e.g., UFD~\citep{ojha2023towards}, Cozzo2024~\citep{cozzolino2024raising}) often lack the representational capacity for diverse distributions, larger models incur prohibitive computational costs. RNSIDNet avoids both extremes by utilizing only 13.58M trainable parameters, ensuring strong generalization with minimal overhead. Computationally, despite its dual-branch architecture, RNSIDNet requires 104.31G FLOPs. This is highly comparable to standard CLIP ViT-L~\citep{radford2021learning} baselines (103G--105G), demonstrating that the introduced noise extraction and modulation components add negligible computational burden. Consequently, RNSIDNet successfully disentangles heterogeneous features without noticeably increasing inference latency, confirming its practicality for real-world applications.

\subsection{Effect of Multi-Source Training Strategy}
\begin{table}[!t]
	\centering
	\caption{Performance comparison on the AIGCDetectionBenchmark~\cite{zhong2023patchcraft} to evaluate the impact of training data sources on cross-architecture generalization. Results are reported in AUC (\%).}
	\label{tab:performance_comparison_auc}
	
	\vspace{0.1em}
	\scriptsize
	\setlength{\tabcolsep}{1.5pt}
	
	\resizebox{\linewidth}{!}{
		\begin{tabular}{lccccccccccccccccc}
			\toprule
			Method & ADM & DALL-E 2 & Glide & Midjourney & VQDM & BigGAN & CycleGAN & GauGAN & ProGAN & SD 1.4 & SD 1.5 & StarGAN & StyleGAN & StyleGAN 2 & WFR & WuKong & AVG \\
			\midrule
			
			RINE~\cite{koutlis2024leveraging} & \textbf{99.25} & \textbf{99.34} & \underline{97.68} & \underline{95.01} & \textbf{99.75} & \underline{99.81} & \underline{99.91} & 99.98 & \textbf{100.00} & \textbf{99.89} & \textbf{99.85} & \underline{99.98} & \underline{98.45} & 88.52 & \textbf{99.99} & \textbf{99.83} & \textbf{98.58} \\
			
			\midrule
			
			RNSIDNet & \underline{96.03} & \underline{99.22} & \textbf{97.85} & \textbf{99.18} & 98.83 & 99.71 & 99.36 & \underline{99.89} & \underline{99.79} & \underline{99.67} & \underline{99.59} & \textbf{99.99} & 96.47 & \underline{93.73} & 83.30 & \underline{99.18} & \underline{97.61} \\

			RNSIDNet (ProGAN) & 93.46 & 86.74 & 94.93 & 89.54 & \underline{98.57} & \textbf{99.97} & \textbf{99.98} & \textbf{100.00} & \textbf{100.00} & 97.47 & 97.37 & \underline{99.98} & \textbf{99.86} & \textbf{99.67} & \underline{99.78} & 98.01 & 97.21\\
            
			\bottomrule
	\end{tabular}}
\end{table}
To investigate the impact of training data sources on model generalization, we evaluate our method under two different training paradigms in Table~\ref{tab:performance_comparison_auc}. Specifically, following the protocol adopted by previous works~\citep{ojha2023towards,koutlis2024leveraging,yan2025a}, we train RNSIDNet exclusively on the ProGAN dataset provided by ~\citet{wang2020cnn}, whereas the standard RNSIDNet is trained on the multi-source AMSID dataset.

As shown in Table~\ref{tab:performance_comparison_auc}, the ProGAN-trained RNSIDNet generalizes exceptionally well across unseen GANs (near 100\% AUC), indicating shared structural artifacts within the same generative family. However, its performance drops significantly on Diffusion models (e.g., 86.74\% on DALL-E 2), exposing a critical cross-architecture domain gap. In contrast, the standard multi-source RNSIDNet successfully bridges this gap, boosting Diffusion model detection to over 99\% while maintaining peak performance on GANs. This validates that relying on a single generator inevitably introduces architecture-specific biases, whereas our multi-source strategy forces the network to capture universal, intrinsic generative artifacts for superior cross-model generalization.

\subsection{Feature Visualization Analysis}
\begin{figure}
	\centering
	\includegraphics[width=1\linewidth, height=0.4\textheight,keepaspectratio]{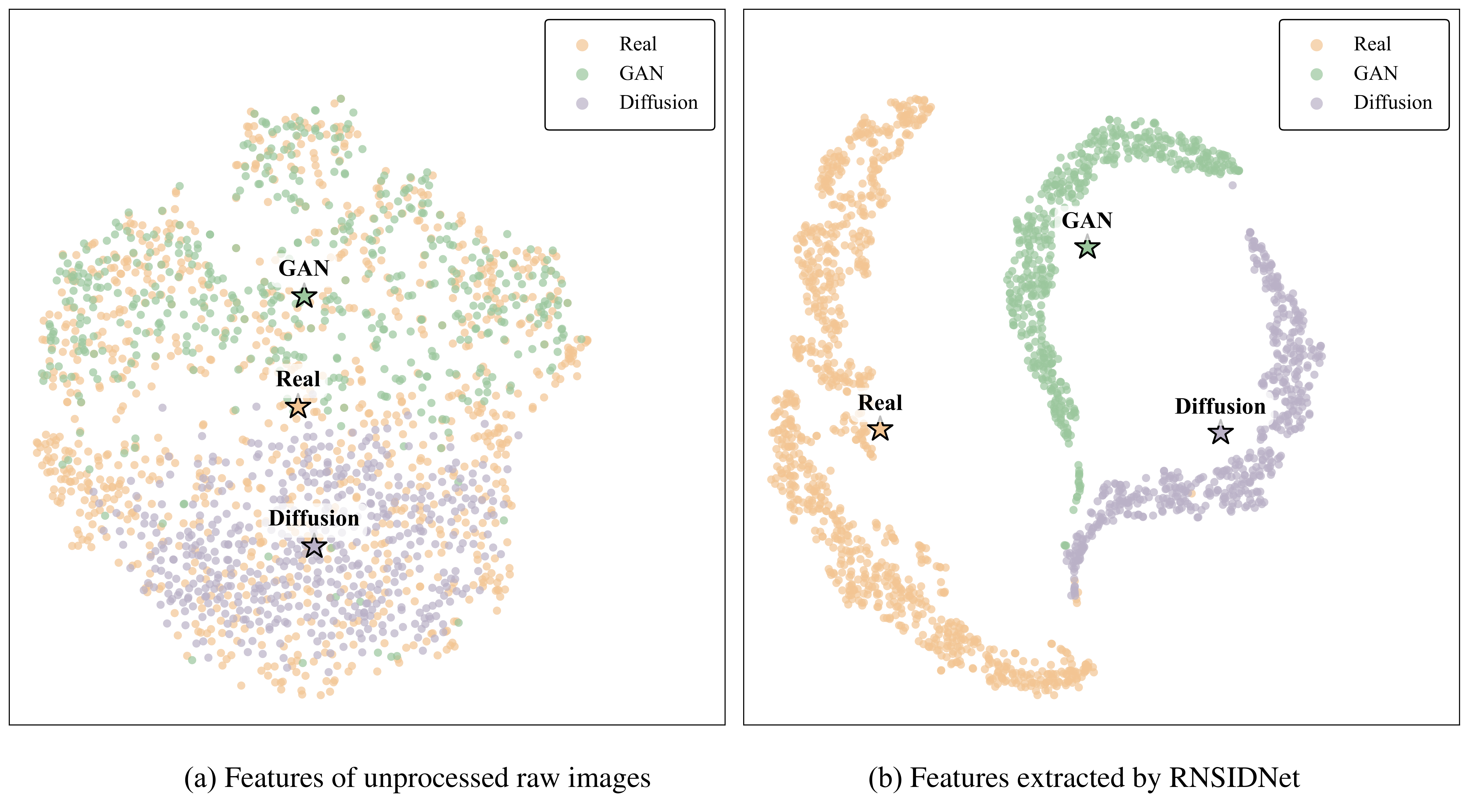} 
	\caption{Feature distribution visualization of real and synthetic images via t-SNE}
	\label{fig:rnsidnet_tsne}
\end{figure}
To evaluate feature disentanglement, we apply t-SNE on equal samples from Real, GAN, and Diffusion categories. As depicted in Fig.~\ref{fig:rnsidnet_tsne}(a), raw image features exhibit severe semantic entanglement. In contrast, RNSIDNet transforms the feature space into an ideal clustering structure, as seen in Fig.~\ref{fig:rnsidnet_tsne}(b). A substantial margin clearly separates pristine and synthetic clusters, validating the efficacy of our dual-branch feature fusion mechanism.

Crucially, GAN and Diffusion samples are further repelled into distinct clusters within the synthetic manifold. This high separability proves that our HSCL strategy successfully enforces strict intra-class compactness and inter-class separation. By successfully decoupling the unique physical fingerprints of heterogeneous generative architectures, RNSIDNet establishes a solid theoretical foundation for its superior cross-domain generalization.

\section{Conclusion}
In this work, we propose RNSIDNet, an innovative dual-branch framework for robust and generalizable synthetic image detection. By synergizing global RGB semantics extracted from a visual foundation model with high-frequency artifacts captured by Bayar-constrained convolutions, our method effectively constructs a comprehensive representation of image authenticity. Crucially, the introduction of our HSCL strategy successfully forces the model to decouple image semantics from inherent generative traces. Comprehensive evaluations validate that RNSIDNet establishes a new state-of-the-art, demonstrating superior cross-model generalization and exceptional resilience against complex post-processing degradations. Moving forward, applying this content-artifact decoupling paradigm to synthetic video forensics remains a promising direction.

\printcredits

\bibliographystyle{cas-model2-names}

\bibliography{ref}

\end{document}